\documentclass[sigconf]{acmart}
\usepackage{listings}
\usepackage{color}
\usepackage{multirow}
\usepackage{bm}
\usepackage{enumitem}
\AtBeginDocument{%
  \providecommand\BibTeX{{%
    \normalfont B\kern-0.5em{\scshape i\kern-0.25em b}\kern-0.8em\TeX}}}

\setcopyright{acmcopyright}
\copyrightyear{2021}
\acmYear{2021}
\setcopyright{acmcopyright}
\acmConference[CIKM '21]{Proceedings of the 30th ACM International Conference on Information and Knowledge Management}{November 1--5, 2021}{Virtual Event, QLD, Australia}
\acmBooktitle{Proceedings of the 30th ACM International Conference on Information and Knowledge Management (CIKM '21), November 1--5, 2021, Virtual Event, QLD, Australia}
\acmPrice{15.00}
\acmDOI{10.1145/3459637.3481972}
\acmISBN{978-1-4503-8446-9/21/11}

\acmSubmissionID{3069}

\begin{document}

\title{PRASEMap: A Probabilistic Reasoning and Semantic Embedding based Knowledge Graph Alignment System}

\author{Zhiyuan Qi}
\authornotemark[1]
\affiliation{%
  \institution{Tencent Jarvis Lab}
  \city{Shenzhen}
  \country{China}
}
\email{qizhyuan@gmail.com}

\author{Ziheng Zhang}
\authornotemark[1]
\affiliation{%
  \institution{Tencent Jarvis Lab}
  \city{Shenzhen}
  \country{China}
}
\email{zihengzhang@tencent.com}

\author{Jiaoyan Chen}
\affiliation{%
  \institution{University of Oxford}
  \city{Oxford}
  \country{UK}
}
\email{jiaoyan.chen@cs.ox.ac.uk}

\author{Xi Chen}
\affiliation{%
  \institution{Tencent Platform and Content Group}
  \city{Shenzhen}
  \country{China}
}
\email{jasonxchen@tencent.com}

\author{Yefeng Zheng}
\authornote{Zhiyuan Qi, Ziheng Zhang, and Yefeng Zheng are also affiliated to Tencent Healthcare (Shenzhen) Co., LTD.}
\affiliation{%
  \institution{Tencent Jarvis Lab}
  \city{Shenzhen}
  \country{China}
}
\email{yefengzheng@tencent.com}

\renewcommand{\shortauthors}{Qi and Zhang, et al.}

\begin{abstract}
    Knowledge Graph (KG) alignment aims at finding equivalent entities and relations (i.e., mappings) between two KGs. The existing approaches utilize either reasoning-based or semantic embedding-based techniques, but few studies explore their combination. In this demonstration, we present PRASEMap, an unsupervised KG alignment system that iteratively computes the \textbf{Map}pings with both \textbf{P}robabilistic \textbf{R}easoning (PR) \textbf{A}nd \textbf{S}emantic \textbf{E}mbedding (SE) techniques. PRASEMap can support various embedding-based KG alignment approaches as the SE module, and it also enables easy human computer interaction that additionally provides an option for users to feed the mapping annotations back to the system for better results. The demonstration showcases these features via a stand-alone Web application with user friendly interfaces. The demo is available at \url{https://prasemap.qizhy.com}.
\end{abstract}





\maketitle

\section{Introduction}
Knowledge Graph (KG) alignment attempts to discover equivalent entities, relations, and attributes (i.e., mappings) between two KGs for constructing larger and more comprehensive KGs.
The previous approaches can be roughly categorized as reasoning-based and embedding-based ones~\cite{ZQW2020VLDB, ZJX2020COLING}.
The traditional reasoning-based approaches, such as PARIS~\cite{FSP2012VLDB} and LogMap~\cite{EB2011ISWC}, often exploit logical reasoning together with lexical and graph matching techniques. However, they are usually weak at utilizing the KG structural information.
In contrast, the embedding-based models, such as GCNAlign~\cite{ZQX2018EMNLP} and MultiKE~\cite{QZW2019IJCAI}, can well explore the graph structures by utilizing KG embedding techniques but they lack holistic analysis or logical consistency check for the discovered mappings~\cite{ZQW2020VLDB}.


In light of the complementarity between the reasoning-based and the embedding-based approaches, we propose to demonstrate PRASEMap, a stand-alone Web application of the KG alignment system PRASE, which is recently proposed in~\cite{qi2021unsupervised}, based on Probabilistic Reasoning (PR) and Semantic Embedding (SE).
As suggested by the name, PRASEMap consists of a PR module, an SE module, and an iterative algorithm for the interaction between two modules~\cite{qi2021unsupervised}.
Its PR module is written in C++ and wrapped in Python, while its SE module can be constructed in Python based on various embedding-based KG alignment algorithms.
It is worth mentioning that PRASEMap provides flexible APIs to allow users to easily develop and evaluate their own SE module implementations.
PRASEMap features several innovative components:

\begin{itemize}
    \item A fully functional knowledge graph alignment system computing/predicting entity mappings across the pre-defined or user-uploaded KG pair.
    \item A registration and login interface helping users initialize, monitor, and resume their assigned task.
    \item An interface visualizing the representative subgraphs of the resultant aligned KG pair and highlighting the mapping relationships.
\end{itemize}

Note that the whole system can be performed in two modes, i.e., an automatic mode without any seed mappings or human intervention and a semi-automatic mode accepting the human annotated mappings for augmentation with an annotation interface specifically designed to identify entity mappings. To summarize, the Web application PRASEMap allows users to test the widely-used benchmarks or upload their own KGs, monitor the execution, annotate uncertain mappings, submit their feedback, and download the resultant mappings. The demonstration of PRASEMap can be experienced online at \url{https://prasemap.qizhy.com}, and its source code is made available at \url{https://github.com/qizhyuan/PRASEMap}.


\section{System Design}

We consider the knowledge graph alignment that aims at discovering the entity mappings, i.e., equivalent entities referring to the same real-world object, between two KGs $\mathcal{KG}_1$ and $\mathcal{KG}_2$.
This section first presents the architecture and pipeline of PRASEMap and then briefly introduces the two core modules --- Probabilistic Reasoning (PR) and Semantic Embedding (SE).

\begin{figure}[t!]
	\centering
	\includegraphics[width=1.0\linewidth]{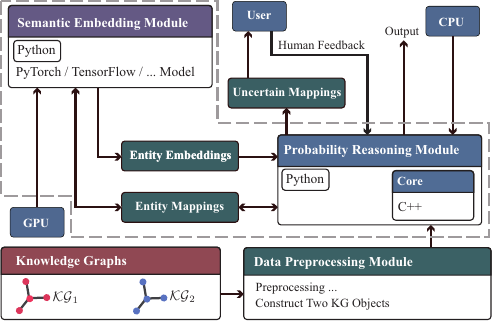}
	\caption{The Architecture of PRASEMap.}
	\label{fig:PRASEMap}
\end{figure}

\subsection{Architecture and Pipeline}

As shown in Figure \ref{fig:PRASEMap}, PRASEMap  consists of the data preprocessing module (denoted as PRE module), the PR module, and the SE module.
These modules work in an iterative pipeline whose steps are described as follow.

\noindent[\textbf{Step 1. Initialization}] The PRE module parses the two KG files that are either pre-stored in the system or uploaded by the user, and constructs two \textit{KG} objects that contain standardized relation and attribute triples. 
Note that for cross-lingual KGs, the PRE module can call the translation API to translate the text into the same language for subsequent PR and SE modules.

\noindent[\textbf{Step 2. Running PR Module}] The PR module receives the \textit{KG} objects, and iteratively discovers cross-KG literal mappings, entity mappings, and relation mappings.
After several self-iterations, it outputs the entity mappings (cf. Section~\ref{pr_module}).

\noindent[\textbf{Step 3. Running SE Module}] The SE module uses the entity mappings from the PR module as seeds for training, and calculates entity embeddings and more entity mappings as the output (cf. Section~\ref{se_module}).

\noindent[\textbf{Step 4. Rerunning PR Module}] The PR module is performed again with the entity embeddings and mappings from the SE module. Different from Step 2, entity embeddings are utilized to enable more accurate probabilistic reasoning, and the received entity mappings are utilized to disclose more potential mappings.  

Given a task, Step 1 and Step 2 are executed once, while Step 3 and Step 4 can be repeated for several times (which can be configured by the user) to fully explore the benefits of the two modules. This is the automatic mode of PRASEMap relying on no seed mapping or human interaction. 
In the semi-automatic mode, the PR module selects and presents some uncertain mappings to the user and receives the manual annotations, based on which it further discovers more mappings and deletes some wrong mappings.



\subsection{Probabilistic Reasoning Module}
\label{pr_module}

\begin{figure*}[!t]
    \centering
    \includegraphics[height=7.5cm]{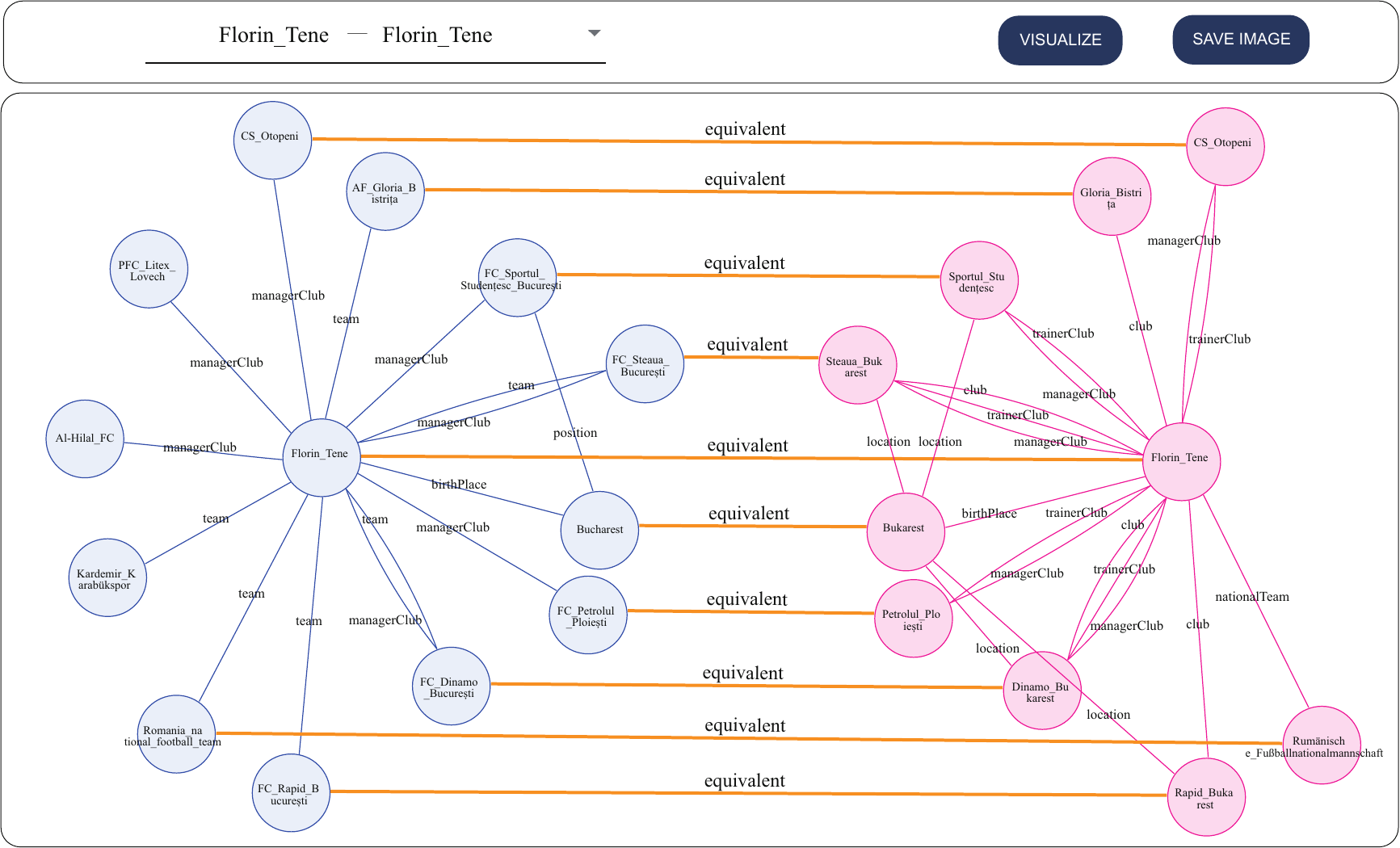}
    \caption{Mapping Visualization.}
    \label{fig:SingleKGDemo}
\end{figure*}

A famous reasoning-based alignment system termed PARIS~\cite{FSP2012VLDB} is adopted and upgraded as the PR module.
PARIS was originally implemented in Java.\footnote{\url{http://webdam.inria.fr/paris/}}
To be compatible with many embedding-based KG alignment algorithms implemented in Python, we re-implement PARIS using C++ and create its Python bindings using the PyBind11 library, which makes PRASEMap much more effective than our original PRASE that is  implemented in Python and evaluated in \cite{qi2021unsupervised}.


Briefly, the original PARIS first computes the functionality of the KG relations and attributes.
Given a head entity, the functionality can be used to determine whether two tail entities or literals across two KGs are identical.
Next, PARIS discovers initial literal mappings and entity mappings by lexical matching, and further iteratively updates the probabilities of these mappings as well as the subsumption relationships between relations and attributes by reasoning with the functionality. 
PARIS finally outputs entity mappings along with other mappings after several iterations when the mappings converge.
Please refer to~\cite{FSP2012VLDB} for more details.

In Step 2, the PR module uses the original PARIS while in Step 4, the PR module uses the upgraded PARIS that has several new features.
First, it can exploit the semantic embeddings output by the SE module to adjust the estimated probability of entity mappings during its reasoning process, so as to improve the recall and the precision of the resultant mappings.
Specifically, the new formulation of computing the equivalent probability of two entities $e_1\in{\mathcal{KG}_1}$ and $e_2\in{\mathcal{KG}_2}$ is given by $\text{Pr}_{\text{new}}(e_1\equiv e_2):=\alpha\text{Pr}_{\text{old}}(e_1\equiv e_2)+(1-\alpha)\text{sim}(\bm{e_1}, \bm{e}_2)$ with $\alpha\in(0,1)$, where the first term corresponds to the original probability estimation formulation of PARIS; the second term measures the (cosine) similarity of entity embeddings used to adjust $\text{Pr}_{\text{old}}$.
We use the Eigen library to manipulate embeddings in the process of reasoning.
Second, the upgraded PARIS directly adopts those high-quality entity mappings calculated by the SE module into its initial entity mapping set for further exploring more potential mappings during its self-iterations.
Please refer to~\cite{qi2021unsupervised} for detailed explanations of the PR module. 

Considering the difficulty in aligning two complex KGs, manual labeling often plays an important role in the real-world applications \cite{ZJX2020COLING}.
To further leverage manual labeling, the PR module enables optional human interaction through which those uncertain mappings are sent to users for feedback.
Specifically, the PR module randomly selects a number of the mappings with low scores for labeling. The users can mark a score of $0$ (i.e., false) or $1$ (i.e., true) for each entity mapping.
Note that PRASEMap regards the labels from user feedback as ground truth and will not update these labeled mappings' scores during the subsequent iterations.


\subsection{Semantic Embedding Module}
\label{se_module}

The default SE module in PRASEMap is constructed based on GCNAlign~\cite{ZQX2018EMNLP}, implemented in TensorFlow.
GCNAlign is a typical embedding-based approach using Graph Convolutional Networks (GCNs) to encode the neighbouring information of entities. It first constructs the adjacency matrices of two KGs based on the functionality of relations, and then embeds entities into a low-dimensional space using GCNs, where equivalent entities are close to each other.

The SE module receives the entity mappings from the PR module and selects the entity mappings with high scores as its training seeds. 
After training, the resultant entity embeddings are used to discover more potential entity mappings by measuring the distance between them. 
The newly discovered mappings and the learned embeddings are fed back to the PR module for iteration (Step 4).

Note that PRASEMap is compatible to different embedding-based KG alignment approaches if they can output entity embeddings and entity mappings.
In addition to GCNAlign, the current PRASEMap system has already implemented MTransE~\cite{MYM2017IJCAI} and MultiKE~\cite{QZW2019IJCAI}, both of which often achieve state-of-the-art performance in many benchmarks.
More SE modules will be added in the follow-up development. 
Users can also easily adopt other embedding-based implementations from external libraries, such as OpenEA~\cite{ZQW2020VLDB}, or develop their own embedding-based KG alignment approaches.

\section{Demonstration}
This section provides a demonstration towards PRASEMap and a brief evaluation of the human annotation during the iterations.

\subsection{Web Application}
For better human-computer interaction, we construct a stand-alone Web application with multiple key features and visualization interfaces.
Specifically, the back end is developed based on SpringBoot, Mybatis, and Redis. The front end is developed using Vue.js and antv/g6 --- a graph visualization engine.
The system itself stores some widely-used KG alignment benchmarks from popular general-purpose KGs, such as DBpedia and Wikidata, as well as an industry benchmark from Tencent's medical KGs~\cite{ZJX2020COLING}. 
The Web application also allows users to upload their own KG files for KG alignment and visualization.
The resultant mappings can be downloaded.
We next introduce three important interfaces and their usage.

\begin{figure}[!t]
    \centering
    \includegraphics[height=7 cm]{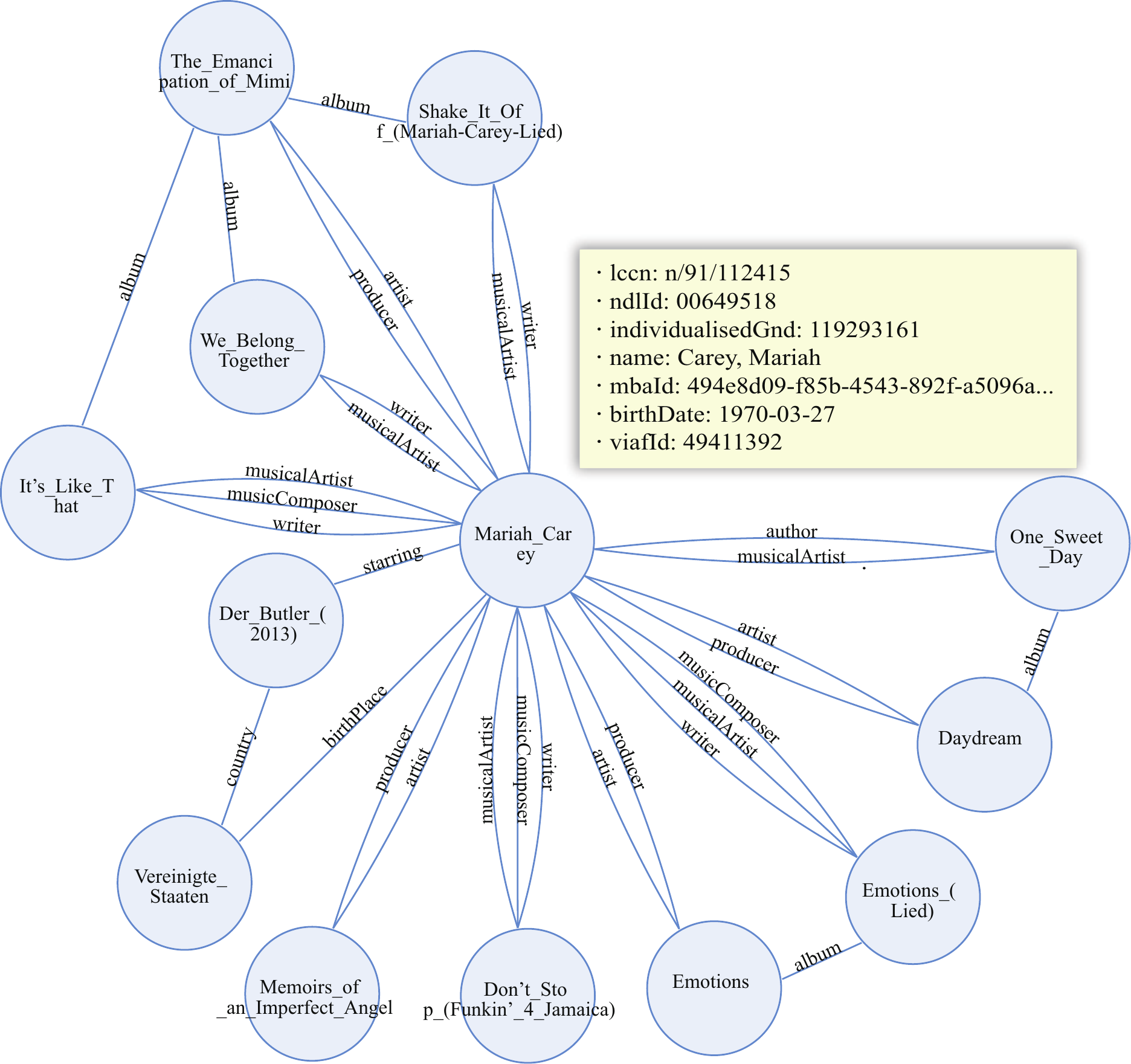}
    \caption{Neighbourhood Visualization.}
    \label{fig:KGMappingDemo}
\end{figure}

\noindent[\textbf{Task Running and Mapping Visualization}]
The user registers at the Web application with his/her email, and then receives a unique task ID for his/her own KG alignment task on the Web application.
After login at the Web application, the user submits a KG alignment task, and the server enqueues the task on a work queue to maintain the workload. 
When PRASEMap starts to run for a task, some logging information is presented for the user to monitor the running process including the running stage, the time cost, the training loss, and the number of discovered mappings.
To illustrate the alignment result, the Web application can visualize each mapping, where the corresponding one-hop neighbourhoods of the two associated entities of the mapping are presented.
With the neighbourhoods, the user can better understand why entities are matched, and meanwhile the Web application can present more mappings between their neighbouring entities, which can further justify why the two entities are matched.
Figure \ref{fig:KGMappingDemo} shows the visualization of an example mapping between ``Florin\_Tene'' of DBpedia-EN (blue-circled entities) and ``Florin\_Tene'' of DBpedia-DE (pink-circled entities),
where the blue/pink lines represent the relations within each single KG, and the orange lines represent all the entity mappings.
Furthermore, the user can click any individual entity with the mouse to check the corresponding attributes that are also helpful to explain a mapping. 

\begin{figure}[t!]
	\centering
	\includegraphics[width=1.0\linewidth]{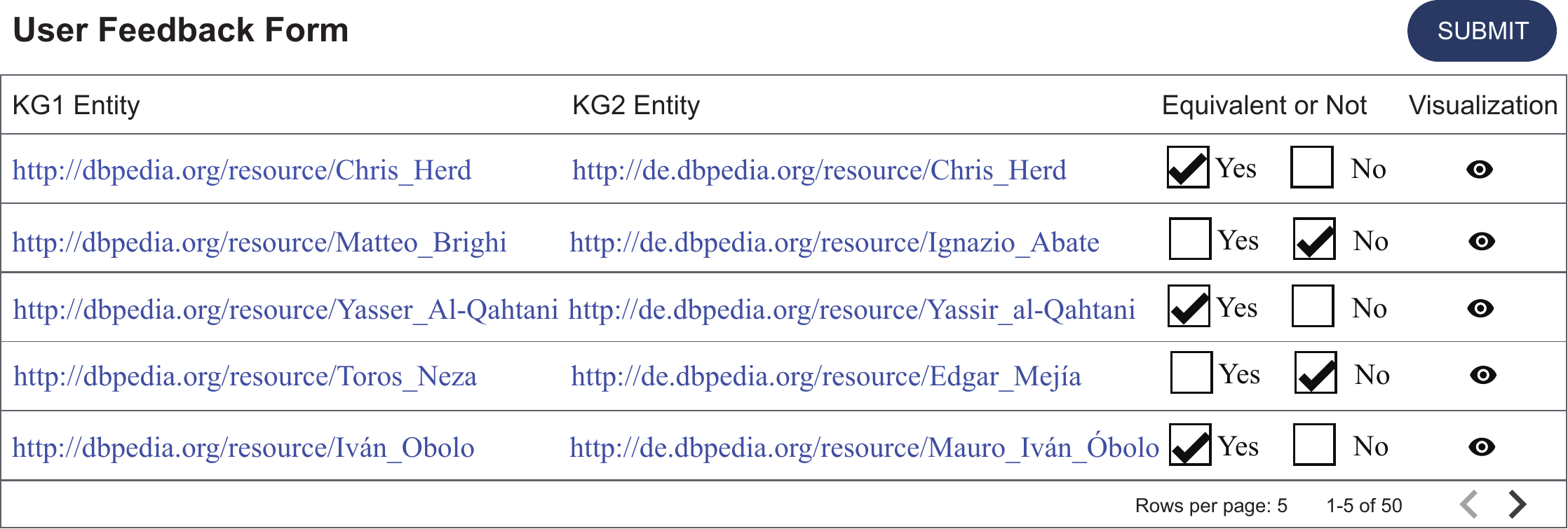}
	\caption{The Interface for Manual Mapping Labeling.}
	\label{fig:HCIDemo}
\end{figure}

\noindent[\textbf{Manual Labeling}] The Web application enables the user to return their feedback to PRASEMap for improvement.
PRASEMap selects a number of entity mappings with low confidence and presents them to the user in a table, named as User Feedback Form in Figure~\ref{fig:HCIDemo}.
For each mapping, the user can judge and annotate whether the two entities are equivalent or not.
To assist manual labeling, another visualization interface is developed to provide the user more information of the involved entities. 
Specifically, when the user clicks the visualization icon, the neighbourhood around each entity (e.g., the connected relations, entities, and attributes) is shown (cf. Figure \ref{fig:SingleKGDemo}).
Manual labeling is fully optional for the user, and the user can label as many mappings as they want. 
Once the user finishes the labeling and submits their feedback, PRASEMap runs again with these user-delivered mappings and updates the alignment results.

\noindent[\textbf{KG and Neighbourhood Visualization}] 
The application can present the overall statistical information of a KG, including the numbers of the entities, relations, and attributes, and the numbers of the relation and attribute triples.
Meanwhile, as aforementioned, the system can visualize the neighbourhood of a specific entity in a KG.
Figure \ref{fig:SingleKGDemo} visualizes the entity ``Mariah\_Carey'' of DBpedia-EN, where its attributes, its one-hop neighbouring entities and connected relations are shown.
The attributes of a neighbouring entity can be displayed when the user clicks it.

\subsection{Evaluation of Human Annotation}


The performance of the automatic mode of PRASEMap has been verified in the performance evaluation of PRASE~\cite{qi2021unsupervised}. This work complements the evaluation on its semi-automatic mode, i.e., the effectiveness of human annotation feedback.

We present the results of PRASEMap on Tencent's industry benchmark MED-BBK-9K~\cite{ZJX2020COLING} using MultiKE as the SE module, with 10, 50 and 100 human annotations fed back to iteration. Note that we use undiscovered ground truth mappings as human feedback for convenience in evaluation.
The results are shown in Table~\ref{tab:prase_performance}, where P, R, and F1 denote precision, recall, and F1-score, respectively.
It can be observed that, compared with PARIS, both the precision and the recall of PRASEMap significantly increase, leading to a higher F1-score.
Besides, the performance of PRASEMap benefits from the human labeling as the P/R/F1 increases as the number of annotations increases.
This finding also sheds light on the importance and effectiveness of human interaction in KG alignment.
Please refer to~\cite{qi2021unsupervised} for more comprehensive evaluation about PRASEMap.


\begin{table}[h]
	\renewcommand\arraystretch{1.05}
	\centering
	\caption{Experimental results on MED-BBK-9K~\cite{ZJX2020COLING}.}
	\setlength{\tabcolsep}{1.2mm}{
		\begin{tabular}{@{}clccc@{}}
			\toprule
			\multicolumn{2}{l|}{Model}                               & \multicolumn{1}{c|}{P (\%)}               & \multicolumn{1}{c|}{R (\%)}                 & \multicolumn{1}{c}{F1 (\%)}               \\ \midrule
			\multicolumn{2}{l|}{PARIS}                             & \multicolumn{1}{c|}{80.588} & \multicolumn{1}{c|}{37.110} & \multicolumn{1}{c}{50.818} \\ \midrule
			\multicolumn{2}{l|}{PRASEMap}  & \multicolumn{1}{c|}{83.833} & \multicolumn{1}{c|}{61.973} & \multicolumn{1}{c}{71.265} \\ \midrule
			\multicolumn{2}{l|}{PRASEMap w/ 10 Annotations} & \multicolumn{1}{c|}{84.186} & \multicolumn{1}{c|}{62.115} & \multicolumn{1}{c}{71.486} \\ \midrule
			\multicolumn{2}{l|}{PRASEMap w/ 50 Annotations} & \multicolumn{1}{c|}{84.402} & \multicolumn{1}{c|}{62.606} & \multicolumn{1}{c}{71.889} \\ \midrule
			\multicolumn{2}{l|}{PRASEMap w/ 100 Annotations} & \multicolumn{1}{c|}{84.760} & \multicolumn{1}{c|}{63.130} & \multicolumn{1}{c}{72.363} \\ \bottomrule
		\end{tabular}
	}
	\label{tab:prase_performance}
\end{table}

\section{Conclusion and Discussion}

In this demonstration, we present a KG alignment system termed PRASEMap that leverages both probabilistic reasoning and semantic embedding for the automatic mapping discovery.
The system is implemented with a Web application with friendly input, output, and visualization interfaces as well as different settings for fully configurable PRASEMap.
It also supports human-computer interaction that utilizes human feedback for better results.

The Tencent team is actively developing new features, such as involvement of ontology \cite{xiang2021ontoea}, and meanwhile maintaining PRASEMap for higher reliability, better performance, and industry deployment.
For future work, we will utilize PRASEMap as a primary component in the construction of large-scale medical KGs.
Moreover, the manual labeling interface in PRASEMap will be integrated into the annotation platform to assist the human annotators and improve their productivity in various tasks beyond mapping annotation, such as KG correction and entity canonicalization~\cite{chen2020correcting}.
We also plan to explore more learning-based strategies for the selection of undiscovered mappings to reduce the annotation workload.

\begin{acks}
This work was funded by National Key R\&D Program of China (2018YFC2000702). We also acknowledged the kind assistance from Shengzhang Lai and Chun Zhang in developing and maintaining the Web application.
\end{acks}

\bibliographystyle{ACM-Reference-Format}
\bibliography{reference}











\end{document}